\documentclass{article}
\usepackage{graphicx} % Required for inserting images

\title{Transformer-Based Representation Learning for Robust Gene Expression Modeling and Cancer Prognosis}

\author{
    Shuai Jiang, PhD\textsuperscript{1} \and
    Saeed Hassanpour, PhD\textsuperscript{1,2,3}
}

\date{
    \textsuperscript{1}Department of Biomedical Data Science, Geisel School of Medicine at Dartmouth, Hanover, NH 03755, USA\\
    \textsuperscript{2}Department of Epidemiology, Geisel School of Medicine at Dartmouth, Hanover, NH 03755, USA\\
    \textsuperscript{3}Department of Computer Science, Dartmouth College, Hanover, NH 03755, USA\\
    E-mails: Shuai.Jiang@dartmouth.edu, Saeed.Hassanpour@dartmouth.edu
}
\usepackage[T1]{fontenc}
\usepackage{cite}
\usepackage{amssymb,amsfonts}
\usepackage{amsmath}
\usepackage{algorithmic}
\usepackage{textcomp}
\usepackage{booktabs}
\usepackage{tabularx}
\usepackage{multirow}
\usepackage{svg}
\usepackage{threeparttable}
\usepackage{hyperref}
\usepackage{xurl}
\usepackage{rotating}
\usepackage{newfloat}
\usepackage{floatrow}
 
\DeclareFloatingEnvironment[name={Supplementary Table}, fileext=lst, listname={List of Supplementary Tables}]{supptable}
\DeclareFloatingEnvironment[name={Supplementary Figure}, fileext=lsf, listname={List of Supplementary Figures}]{suppfigure}
\begin{document}

\maketitle

\begin{abstract}
Transformer-based models have achieved remarkable success in natural language and vision tasks, but their application to gene expression analysis remains limited due to data sparsity, high dimensionality, and missing values. We present GexBERT, a transformer-based autoencoder framework for robust representation learning of gene expression data. GexBERT learns context-aware gene embeddings by pretraining on large-scale transcriptomic profiles with a masking and restoration objective that captures co-expression relationships among thousands of genes. We evaluate GexBERT across three critical tasks in cancer research: pan-cancer classification, cancer-specific survival prediction, and missing value imputation. GexBERT achieves state-of-the-art classification accuracy from limited gene subsets, improves survival prediction by restoring expression of prognostic anchor genes, and outperforms conventional imputation methods under high missingness. Furthermore, its attention-based interpretability reveals biologically meaningful gene patterns across cancer types. These findings demonstrate the utility of GexBERT as a scalable and effective tool for gene expression modeling, with translational potential in settings where gene coverage is limited or incomplete.
\end{abstract}


\section{Introduction}

Gene expression profiling provides a powerful window into the molecular landscape of cancer, enabling a better understanding of tumor heterogeneity, disease progression, and therapeutic response~\cite{Wang2005Gene-expression,Zhang1997Gene}. Changes in the expression of genes that regulate cell proliferation and differentiation play a central role in oncogenesis~\cite{Bishop1987molecular}, and specific expression signatures have been identified as prognostic biomarkers for survival and recurrence across multiple cancer types~\cite{Tian2010Biological,Pu2020Research-based}. These transcriptomic features can guide risk stratification, inform treatment decisions, and potentially improve clinical outcomes~\cite{Dubsky2020clinical,Chang2003Gene,Boland2021Value}.

Advances in high-throughput technologies such as RNA sequencing (RNA-seq) have made it possible to profile the transcriptome at scale and with high precision~\cite{Dijk2014Ten}. Together with microarray platforms~\cite{Mantione2014Comparing,Lai2020Overall}, these technologies have contributed to large public datasets like TCGA and GEO, facilitating broad access to transcriptomic profiles across diverse cancer types. At the same time, machine learning methods have emerged as powerful tools for extracting predictive patterns from gene expression data, enabling models to capture complex, nonlinear interactions among genes~\cite{Alharbi2023Machine,Kourou2015Machine,Cruz2006Applications}. Nonetheless, the high dimensionality of gene expression data, combined with relatively small sample sizes and frequent missingness, presents ongoing challenges. Traditional dimensionality reduction or feature selection techniques~\cite{Tibshirani1996Regression,Wu2019Selective,Khademi2015Probabilistic} often discard contextual gene relationships and fail to generalize across tasks.

Transformer models have transformed natural language processing by leveraging multi-head self-attention to capture contextual dependencies across sequences~\cite{Vaswani2017}. Pretrained language models such as BERT and GPT have demonstrated the power of transfer learning in data-limited settings~\cite{Devlin2019,RadfordImproving}, offering a promising paradigm for biomedical data modeling. In the transcriptomic context, genes can be treated as symbolic units analogous to words, with their expression patterns reflecting latent biological states. This analogy has inspired recent efforts to adopt transformers for modeling gene expression~\cite{Yang2022scBERT,Khan2023DeepGene,Zhang2022Transformer}, although most prior work has focused on classification tasks or single-cell analysis, often without addressing real-world issues like missing values or the need for generalizable representations.

In this study, we introduce GexBERT, a transformer-based autoencoder framework for representation learning on bulk RNA-seq data. GexBERT is pretrained using a masking and restoration objective that encourages the model to learn distributed representations of gene expression patterns by predicting masked values from context genes. This self-supervised pretraining approach allows GexBERT to capture co-expression structure and functional dependencies among genes without requiring phenotype labels. To evaluate its effectiveness, we assess GexBERT across three downstream applications central to cancer genomics: pan-cancer classification, cancer-specific survival prediction, and missing value imputation. Our results demonstrate that GexBERT achieves strong predictive performance from limited gene subsets, outperforms classical imputation methods, and provides biologically meaningful attention patterns, highlighting its potential as a scalable and interpretable tool for gene expression modeling in translational oncology.

\section{Related work}
\subsection{Deep learning approaches for gene expression-based survival prediction}
Recent research in the field of cancer survival prediction using RNA sequencing data has witnessed significant advancements driven by deep learning techniques. For example, DeepSurv and Cox-nnet used a multi-layer artificial neural network (ANN) model to encode RNA-seq data for survival analysis\cite{Ching2018Cox-nnet:,Katzman2018DeepSurv:}. AECOX integrates an autoencoder with a Cox regression network to simultaneously learn a lower dimensional representation of the input data and predict survival outcomes \cite{Huang2020Deep}. In another study for GBM cancer patients, gene expression data are used as inputs to build a deep multilayer perceptron network to predict patient survival risk. Genes that are important to the model are identified by the input permutation method \cite{Wong2019Prognostic}. In additional to using the deep learning architecture to model the interactions of input gene expression, transfer learning approach have also been explored by pre-training the model using unlabeled dataset and it was found to be effective especially for small training dataset\cite{Hanczar2022Assessment}.

Convolutional Neural Network (CNN) methods have also been used in modeling gene expression data. This group of methods usually first transform the one-dimensional expression data into two-dimensional matrices, then use CNN layers to extract higher-level features. For example, Lopez-Garcia et al. transformed RNA-seq samples into gene-expression images, and pre-trained CNN architectures for survival prediction. Then the model are fine-tuned for lung cancer progression-free interval prediction\cite{Lopez-Garcia2020Transfer}. The CNN-Cox model combines a special CNN framework with prognosis-related feature selection to reduce computational demand\cite{Yin2022convolutional}. Wang et. al. converted gene expression data into two kinds of images with KEGG BRITE and KEGG Pathway data incorporated and then used CNN model to learn high-level features \cite{Wang2022Novel}. REFINED-CNN maps high-dimensional RNA sequencing data into REFINED images which are conducive to CNN modeling. This approach can be adapted to new downstream tasks using transfer learning\cite{Bazgir2023REFINED-CNN}.

\subsection{Distributed embedding}
% Develop a novel deep learning approach to extract prognostic features from multi-omics data based on semantic embeddings. 
In language models, words are converted into dense vectors (i.e., semantic embeddings) through the use of techniques such as word2vec or Bidirectional Encoder Representations from Transformers (BERT) trained on large corpora (e.g., Wikipedia)\cite{Devlin2019, Mikolov2013Efficient}. These embeddings can capture complex word semantics and greatly benefit downstream tasks, especially those with a small sample size \cite{Howard2018Universal}. Similarly, genomic profiles can be viewed as information in a gene “alphabet”, and thus benefit from the latest language modeling techniques. 

Some previous work has been proposed for multi-omics data modeling, including ProtVec (and its generalization BioVec), gene2vec, and mut2vec. ProtVec was trained to project proteins to a continuous space using n-gram models for amino acid sequence\cite{Asgari2015Continuous}. Gene2vec is based on the co-expression of gene pairs\cite{Du2018Gene2Vec:}, and mut2vec is based on the co-occurrence of mutated genes in a randomly sampled bag of genes\cite{Kim2018Mut2Vec:}. These studies yielded promising results; for example, the gene2vec study showed that genes in the known pathways tend to have higher cosine similarity than a random set of genes. Despite the promising implications of these studies, it is methodologically difficult to directly apply them to the prognosis prediction task based on multi-omics data. This is because these genomic biomarkers are continuous variables, while the current modeling methods are limited to handling binary variables (i.e., whether each gene is mutated or not).

\subsection{Transformer-based gene expression modeling}

With the introduction of transformer models, emerging studies have also explored the application of transformers into modeling the gene expression data. T-GEM, or Transformer for Gene Expression Modeling, is a deep learning model based on Transformer that can predict gene expression-based phenotypes. It uses self-attention to model gene-gene interactions and learn new representations of input genes. T-GEM used gene dependent weights to project gene representation in QKV projections, and was shown to capture important biological functions and marker genes associated with the predicted phenotypes\cite{Zhang2022Transformer}. scBERT is a transformer-based model for cell type annotation of single-cell RNA-seq data. It is inspired by BERT and consists of two phases: pre-training and fine-tuning. In the pre-training phase, scBERT learns the gene-gene interaction from a large amount of unlabeled scRNA-seq data using a masked language modeling objective. In the fine-tuning phase, scBERT adapts to the specific cell annotation task of a given scRNA-seq dataset using a supervised classification objective \cite{Yang2022scBERT}. DeepGene Transformer uses a hybrid of one-dimensional convolutional layers prior to the application of multi-head self-attention module to reduce the length of transformer input \cite{Khan2023DeepGene}. 

\subsection{Missing data imputation}

Gene expression data also prone to missing data due to various reasons that occur during the multi-step procedure in data preparation and measurements. Previous studies have found that gene expression data might have up to 5\% of missing values\cite{Chiu2013Missing, Brevern2004Influence}. Common missing data imputation approach include mean/median replacement, K-Nearest Neighbors (KNN) imputation\cite{JSSv074i07}, and Multiple Imputation by Chained Equations (mice)\cite{Buuren2011mice:}. In a previous study, the impact of five well-known missing value imputation methods, including Bayesian Principal Component Analysis (BPCA), KNN, local least squares (LLS), Mean and Median replacement, on clustering and classification tasks were evaluated using 12 cancer gene expression datasets. The study finds that simple methods like mean or median imputation perform as well as more complex strategies\cite{Souto2015Impact}. In another study, Zhu et al. proposes an ensemble method that combines multiple single imputation methods to handle missing values in genomics datasets\cite{Zhu2021efficient}. Lastly, for situations where the entire expression profile of one patient is missing, Zhou et al. introduces TDimpute that predicts missing gene expression data from DNA methylation data. The model is trained on a pan-cancer dataset and fine-tuned for specific cancer types\cite{Zhou2020Imputing}.

\section{Methods} 
\subsection{Transformer-based auto-encoder for gene expression modeling} 

\begin{figure}
    \centering
    \includegraphics[width=\textwidth]{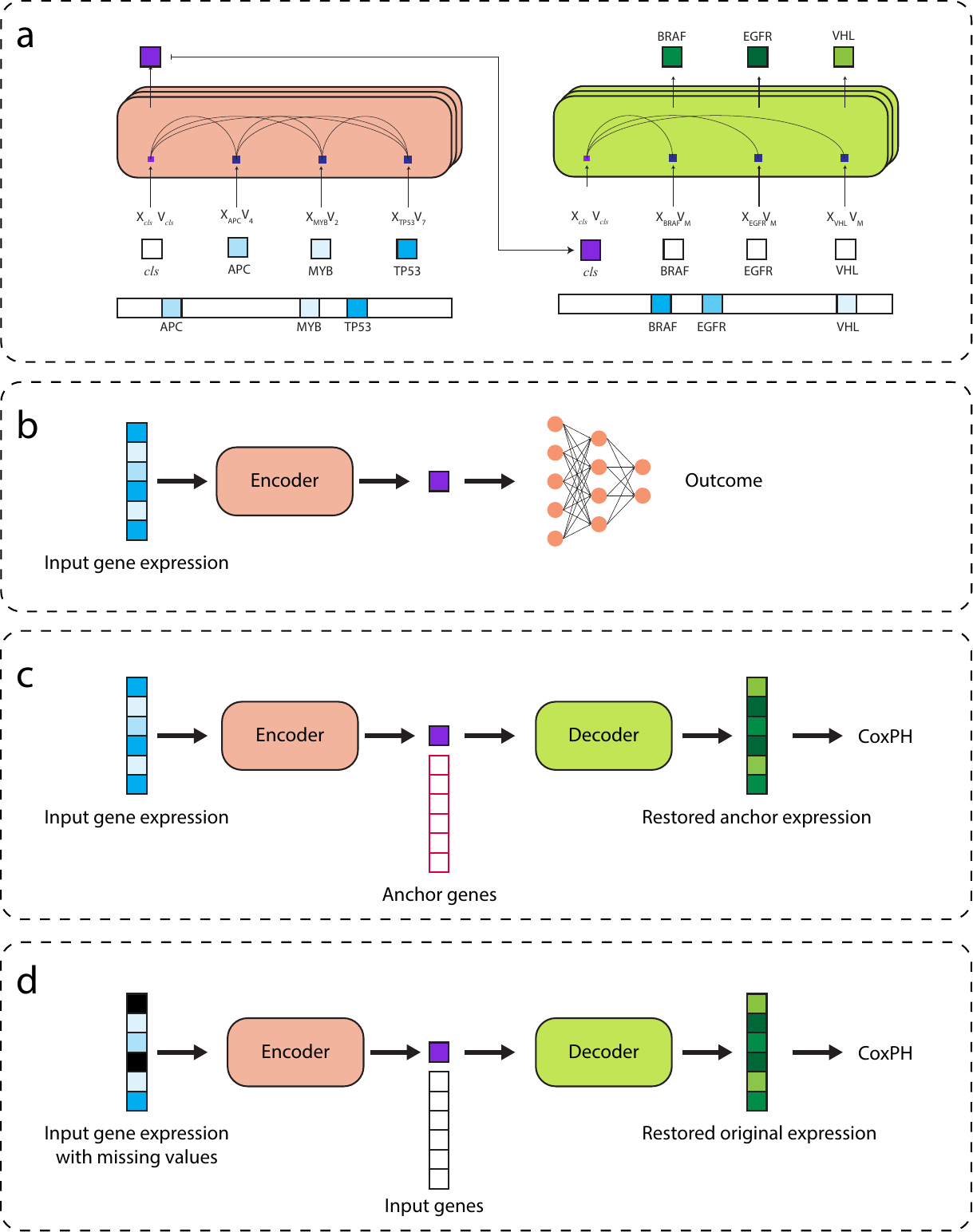}
    \caption{Overview of the GexBERT framework. a) Pretraining: The transformer learns gene co-expression patterns by encoding a subset of input genes and restoring another subset. b) Cancer classification: The encoder extracts gene expression representations for tumor type prediction.
(c) Survival prediction: The model restores prognostic anchor gene expression, which is used in a Cox proportional hazards (CoxPH) model. d) Missing value imputation: The model imputes missing gene expression values to enhance downstream survival prediction.}\label{fig:rna-overview}
\end{figure}

To learn a distributed representation of genetic symbols, we developed transformer-based representation framework named gene expression BERT (GexBERT). While the general structure of BERT has been well documented\cite{Devlin2019}, it cannot be directly applied to the genes, due to the continuous expression values that are associated with each gene. Our modifications and adaptations are described below and are shown in \autoref{fig:rna-overview}a. Briefly, GexBERT consists of a transformer encoder and transformer decoder. The encoder randomly selects a set of genes, and assigns gene embedding and value embedding to each gene and its associated expression value. A class token (\textit{cls}) is appended to the input sequence, which is then processed by the transformer encoder. The summary embedding (i.e., \textit{cls} token) from the encoder output is then used to decode the expression value of another set of genes sampled from the same patient. At this point, only the gene embeddings are kept, and the decoder needs to predict the expression value of each gene using the provided summary embedding.

\subsection{Random sampling of biomarkers}

The computational complexity of the transformer model scales quadratically with the input length. A sentence usually consists of less than 100 words, while the human genome has more than 20,000 genes. As such, it is infeasible to use all genes during each training iteration to learn their embeddings all at once. Instead, in each iteration, we adopt a random sampling approach to obtain $N_{gene_i}$ genes for training. This approach aligns with similar strategies employed in studies like mut2vec \cite{Kim2018Mut2Vec:}. Notably, in contrast to the gene2vec study, which focused on the co-expression of individual gene pairs, our GexBERT approach incorporates a larger context size, enabling the modeling of more intricate interactions among genes.

\subsection{Genetic symbol and value encoding}

% As the arrangements of genes lacks meaningful syntactic information, we opt to replace positional encoding with gene expression value encoding to signify the status of each gene. In handling the continuous values of expression level, we discretize them into categorical numbers of $N_{levels}$. This discretization enables us to use $N_{levels}$ learnable embedding vectors to represent the expression information of genes. In instances where measurements for any biomarker are missing, a special learnable token (\textit{missing}) is employed. As for the genetic symbols, we randomly initialize an embedding vector for each gene $X_{gene}\in R^{d\times1}$. 

Since the arrangement of genes lacks inherent syntactic structure, we replace positional encoding with gene expression value encoding to represent the state of each gene. Given that gene expression levels are continuous, we discretize them into \( N_{\text{levels}} \) categorical bins, where each level \( j \) is associated with a learnable embedding vector \( \mathbf{e}_{j} \in \mathbb{R}^{d} \). This results in an embedding matrix \( \mathbf{E}_{\text{expr}} \in \mathbb{R}^{N_{\text{levels}} \times d} \), which captures expression information. For genes with missing expression measurements, we introduce a special learnable token \( \mathbf{e}_{\text{missing}} \in \mathbb{R}^{d} \).  

Each gene \( i \) is assigned a randomly initialized embedding vector \( \mathbf{x}_{i} \in \mathbb{R}^{d} \), and its final representation is obtained by summing the gene embedding and its corresponding expression embedding:  

\[
\mathbf{x}_{i}^{(j)} = \mathbf{x}_{i} + \mathbf{e}_{j}
\]

where \( \mathbf{x}_{i}^{(j)} \) denotes the combined embedding of gene \( i \) with discretized expression level \( j \).

\subsection{Autoencoder architecture}

% Our pre-training model comprises two key components: a transformer encoder and a transformer decoder. During the encoding step, a subset of genes, denoted as $S_{input}$, is randomly sampled for each patient. The gene and value embedding method is then applied to convert each gene and its associated expression value into corresponding embeddings. These embeddings, along with the learnable class token (\textit{cls}), are input into the transformer encoder. The transformer encoder comprises $n$ repetitions of $h$ multi-head self-attention heads.
Our pre-training model consists of two primary components: a transformer encoder and a transformer decoder. During the encoding phase, a subset of genes, denoted as $S_{\text{input}}$, is randomly sampled for each patient. The gene expression encoding method is then applied, where each gene and its corresponding expression value are transformed into learnable embeddings. These embeddings, along with a learnable classification token (\textit{cls}), serve as inputs to the transformer encoder. The encoder comprises $n$ layers, each containing $h$ multi-head self-attention mechanisms, enabling contextualized representation learning of gene interactions.

% During the restoration step, we either utilize the same set of input genes or randomly sample another $N_{gene_o}$ genes from the same patient, contingent upon the pre-training and evaluation strategy employed. Subsequently, the expression values corresponding to these genes are masked, employing a common mask token to denote the masking information. The gene embedding and the masking embedding are concatenated with the summary embedding from the encoding step, forming the input for the decoder transformer.
During the restoration phase, we either use the same set of input genes or randomly sample an additional $N_{\text{gene}_o}$ genes from the same patient, depending on the pre-training and evaluation strategy. The expression values of these selected genes are then masked using a shared masking token to indicate missing information. The corresponding gene embeddings and masking embeddings are concatenated with the summary embedding from the encoding step, forming the input to the transformer decoder.

% The decoder transformer shares the same structure as the encoder transformer, with the exception that we disable self-attention among the masked genes. Instead, attention is solely permitted between the masked genes and the class token. This modification is made because the selection of genes to decode should not impact the restored expression levels of other genes. Moreover, this alteration significantly expedites the decoding process by replacing the costly MHSA with operations that scale linearly with the length of genes to be restored.
The transformer decoder follows the same architectural structure as the encoder, except that self-attention among masked genes is disabled. Instead, attention is restricted to interactions between the masked genes and the class token. This design choice ensures that the selection of genes for decoding does not influence the restored expression levels of other genes. Additionally, this modification significantly accelerates the decoding process by replacing MHSA with operations that scale linearly with the number of genes to be restored.

From the decoder transformer's output, we apply a linear projection layer to map the embeddings of genes to be restored into their predicted expression values. The training objective is to minimize the restoration loss, quantified by the mean squared error (MSE):  

\[
MSE = \frac{1}{n} \sum_{i=1}^{n} (y_i - \hat{y}_i)^2
\]

where \( y_i \) denotes the true expression value, \( \hat{y}_i \) represents the predicted value, and \( n \) is the number of genes to be restored. The pre-training process aims to capture intricate gene interactions by optimizing the model to accurately reconstruct the expression values of randomly selected genes.

\section{Experiments} 

\subsection{Dataset}

The dataset utilized in this study was extracted from the gene expression panel of The Cancer Genome Atlas (TCGA), obtained from Xena (\url{https://xena.ucsc.edu}). The gene expression dataset encompasses all 33 cancer types, with gene expression values normalized to ensure a mean of zero and a standard deviation of one. After filtering out genes with minimal variations, the final dataset comprised a total of 11,014 patients and 20,311 genes.

A comprehensive description of the dataset is provided in \autoref{table:rna-data}. The evaluation specifically focused on 14 cancer types with at least 100 events (i.e., patients who died during follow-up). 30\% of patients from these cancer types were included as the test split. The remaining 70\% of patients from these cancer types, along with all patients from other cancer types, constituted the development split. 80\% of the development dataset was used for pre-training the transformer model, and the rest 20\% of the development dataset was reserved for monitoring the loss during training. 

\begin{table}[]
\centering
\caption{Summary of datasets and data splits. Number of patients and observed events (deaths) in the development (Dev) and test splits across cancer types used in the experiments.``Others'' includes cancer types with fewer than 100 observed events, used only in pretraining.}
\label{table:rna-data}

\begin{tabular}{@{}lcccc@{}}
\toprule
\multicolumn{1}{c}{\multirow{2}{*}{Cancer}} & \multicolumn{2}{c}{Dev Split} & \multicolumn{2}{c}{Test Split} \\ \cmidrule(l){2-5} 
\multicolumn{1}{c}{}                        & Patients       & Events       & Patients        & Events       \\ \cmidrule(l){2-5} 
BLCA                                        & 320            & 142          & 106             & 47           \\
BRCA                                        & 910            & 148          & 304             & 50           \\
COAD                                        & 368            & 85           & 123             & 28           \\
GBM                                         & 125            & 100          & 41              & 33           \\
HNSC                                        & 424            & 190          & 141             & 63           \\
KIRC                                        & 454            & 151          & 152             & 51           \\
LAML                                        & 121            & 77           & 40              & 26           \\
LGG                                         & 396            & 100          & 132             & 33           \\
LIHC                                        & 317            & 123          & 105             & 41           \\
LUAD                                        & 425            & 157          & 142             & 52           \\
LUSC                                        & 409            & 183          & 137             & 61           \\
OV                                          & 230            & 139          & 77              & 46           \\
SKCM                                        & 343            & 161          & 115             & 54           \\
STAD                                        & 333            & 125          & 110             & 41           \\
Others                                      & 4114           & 801          & -               & -            \\ \bottomrule
\end{tabular}
\end{table}

\subsection{Model pre-training}
In each iteration, we independently and randomly sample 512 genes for each patient, converting their normalized expression values into one of 64 value tokens ($N_{\text{levels}}$=64). The embedding dimension is set to 200 for both gene embeddings and value embeddings. Subsequently, a transformer encoder with 8 heads and 8 layers is used, and the output of the summary embedding is obtained for restoration. During the restoration phase, the same set of genes used as the input set is utilized for restoration. This approach resembles the variation autoencoder type of pre-training, with the distinction that the set of genes for each patient and during each iteration varies and is randomly sampled. 

The transformer model was implemented in Pytorch and optimized using AdamW. The pre-training was performed on 4 NVIDIA RTX 6000 GPUs, and each epoch required approximately 15 seconds. Following 10,000 epochs of pre-training, the restoration loss reduced to 0.160. Subsequently, the model was further pre-trained for an additional 1,500 epochs using randomly sampled genes for restoration. The final model is capable of restoring randomly sampled genes with a MSE loss of 0.272.

\subsection{Evaluation of the pre-trained GexBERT model for downstream tasks}

We evaluated the pre-trained GexBERT model with three types of downstream tasks. To obtain more stable estimate of the performance metrics for each task, we repeated each experiment 10 times and reported the average performance measurements.

\subsubsection{Pan-cancer classification}
To assess GexBERT's performance in pan-cancer classification (\autoref{fig:rna-overview}b), we conducted two series of experiments. In the first series, we extracted the summary embedding from GexBERT's output and trained a linear layer to predict the type of cancer based on a varying number of randomly selected genes, ranging from 64 to 1024. We compared the performance of GexBERT's summary embedding with that of the top 200 components extracted from Principal Component Analysis (PCA). Furthermore, while continuing to utilize the summary embedding for pan-cancer classification, we fine-tuned the transformer backend of GexBERT. For comparison, we evaluated its performance against a 2-layer neural network model with 128 hidden units, while both models using the same set of randomly selected genes as input. Accuracy was used as the performance metric for comparison.

\subsubsection{Cancer-specific survival prediction}

For the 14 cancer types with the highest number of events during follow-up, we evaluated the performance of the GexBERT model in cancer-specific survival prediction (\autoref{fig:rna-overview}c). Unlike pan-cancer classification, cancer-specific survival prediction is a more challenging task as it focuses on the differential expression of genes among patients within the same cancer type.  

To better adapt the GexBERT model for survival prediction, we identified a set of anchor genes for each cancer type. These anchor genes exhibit high prognostic significance and were selected using univariate Cox Proportional Hazards (CoxPH) analysis. Specifically, we assessed the survival predictive power of each gene within the development dataset through cross-validation, evaluating its performance for each cancer type. Genes that achieved the highest cross-validated concordance index (C-index) were selected as anchor genes.  

During evaluation, we used various gene sets as inputs to the GexBERT model and used it to restore the expression levels of the selected anchor genes. After obtaining the restored gene expression values, we directly applied them for survival prediction using an $L_2$-regularized Cox Proportional Hazards (CoxPH) model. We cross-validated the cancer-specific C-index of the restored expression values in the test split and compared it to that of the original expression values from the input genes.  

Additionally, we evaluated the prognostic ability of the summary embedding that was extracted from the pre-trained GexBERT model.

\subsubsection{Missing data imputation}

We conducted an evaluation of GexBERT's performance in scenarios where missing values were present. To simulate Missing Completely At Random (MCAR) situations, we introduced missing rates ranging from 1\% to 50\%. In MCAR scenarios, each element in the input dataset has an equal chance of being missing. GexBERT inherently handles missing values by creating an attention mask to discard the incorporation of the missing tokens in the encoder transformer, ensuring they did not contribute to the encoding process.

During evaluation, we utilized the GexBERT model with missing inputs to impute the expression value of the input genes (\autoref{fig:rna-overview}d). The restored expression values were then utilized for survival analysis, following the methodology described in the previous section.

In this evaluation, we compared GexBERT's performance to several commonly used approaches for handling missing values, including:
\begin{itemize}
    \item Zero imputation: Replacing missing values with 0s.
    \item Average imputation: Replacing missing values with the column average.
    \item K-Nearest Neighbors imputation.
    \item Multiple Imputation by Chained Equations (MICE).
\end{itemize}

This comparison enabled us to evaluate GexBERT's effectiveness in handling missing values and its performance relative to conventional imputation methods. Given the high computational cost of MICE, we included it as a benchmark only when there were 64 predictors.

\section{Results}

\subsection{Pan-cancer Classification}

We first evaluated the accuracy of pan-cancer classification by comparing GexBERT with the PCA method and a NN model. The results, summarized in \autoref{table:rna-classification-comp}, demonstrate that the summary embedding of GexBERT consistently outperforms the top 200 PCA components in terms of classification accuracy. When comparing the NN model with the fine-tuned GexBERT model, we observe that GexBERT achieves superior performance with up to 256 genes, with the most pronounced improvement occurring at 64 genes (0.940 vs. 0.928). However, the benefits of pre-training plateau when the number of input genes exceeds 512.

\begin{figure}
    \centering
    \includegraphics[width=\textwidth]{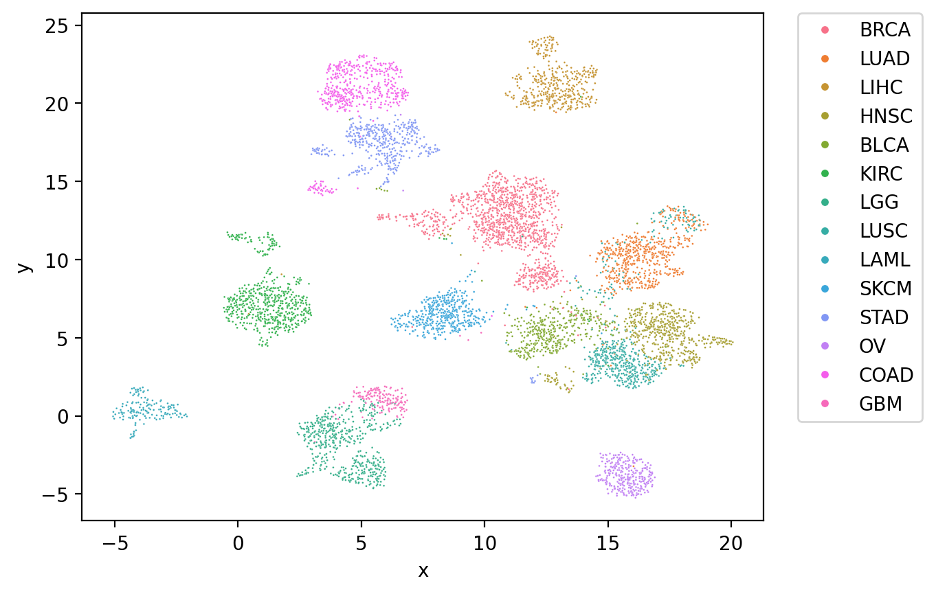}
    \caption{UMAP visualization of GexBERT-generated embeddings. Low-dimensional projection of summary embeddings extracted from GexBERT with 4,096 input genes, showing distinct clustering patterns across cancer types.}\label{fig:rna-cancer-umap}
\end{figure}

\begin{table}[]
\centering
\caption{Pan-cancer classification performance across different gene set sizes. Mean accuracy (standard deviation) for classification using PCA-200, GexBERT (\textit{cls} token embedding), a neural network (NN) baseline, and fine-tuned GexBERT. Results demonstrate that GexBERT consistently outperforms PCA-based embeddings and achieves competitive performance with the NN model, particularly when fine-tuned.}
\label{table:rna-classification-comp}
\begin{tabular}{@{}lcc|cc@{}}
\toprule
Number of genes & PCA-200       & GexBERT & NN-model      & GexBERT fine-tuned \\ \midrule
64              & -             & 0.927 (0.014)   & 0.928 (0.014) & 0.940 (0.011)       \\
128             & -             & 0.958 (0.004)   & 0.955 (0.005) & 0.962 (0.005)       \\
256             & 0.968 (0.004) & 0.969 (0.005)   & 0.969 (0.004) & 0.971 (0.005)       \\
512             & 0.974 (0.002) & 0.975 (0.003)   & 0.975 (0.003) & 0.975 (0.003)       \\
1024            & 0.975 (0.002) & 0.977 (0.002)   & 0.979 (0.002) & 0.979 (0.002)       \\
% 4096            & 0.977 (0.003) & 0.978 (0.001)   & 0.983 (0.001) & 0.983 (0.001)       \\ 
\bottomrule
\end{tabular}
\end{table}

\autoref{fig:rna-cancer-umap} presents the UMAP visualization of the summary embedding output from GexBERT using 4,096 input genes. The figure demonstrates that the summary embedding effectively captures cancer-type distinctions, effectively capturing distinct and coherent clusters.

\subsection{Survival prediction}
% \begin{figure}
%     \centering
%     \includegraphics[width=\textwidth]{survival-pca-vs-cls.png}
%     \caption{Comparison of PCA (up to 200 components) and summary embedding for cancer specific survival prediction measured in average C-index}\label{fig:rna-survival-cls-pca}
% \end{figure}

\begin{table}[]
\centering
\caption{Survival prediction performance across different gene set sizes.
Mean concordance index (C-index $\pm$ standard deviation) for survival prediction using PCA-based embeddings and GexBERT (\textit{cls} token embedding). GexBERT consistently outperforms PCA, demonstrating its ability to capture prognostic signals from gene expression data more effectively.}
\label{table:rna-survival-pca-cls}
% \begin{tabular}{@{}lcc|cc@{}}
\begin{tabular}{lcc}
\toprule
\multicolumn{1}{l}{Input genes} & PCA         & GexBERT  \\ \midrule
64                              & -           & 0.606±0.011  \\
128                             & -           & 0.614±0.012  \\
256                             & 0.559±0.013 & 0.618±0.006  \\
512                             & 0.568±0.011 & 0.625±0.008  \\
1024                            & 0.564±0.009 & 0.625±0.007  \\
\bottomrule
\end{tabular}
\end{table}

% \begin{table}[]
% \begin{tabular}{rll}
% \multicolumn{1}{l}{Input genes} & pca         & \textit{cls} \\
% 64                              & -           & 0.606±0.011  \\
% 128                             & -           & 0.614±0.012  \\
% 256                             & 0.559±0.013 & 0.618±0.006  \\
% 512                             & 0.568±0.011 & 0.625±0.008  \\
% 1024                            & 0.564±0.009 & 0.625±0.007 
% \end{tabular}
% \end{table}

\begin{figure}
    \centering
    \includegraphics[width=0.7\textwidth]{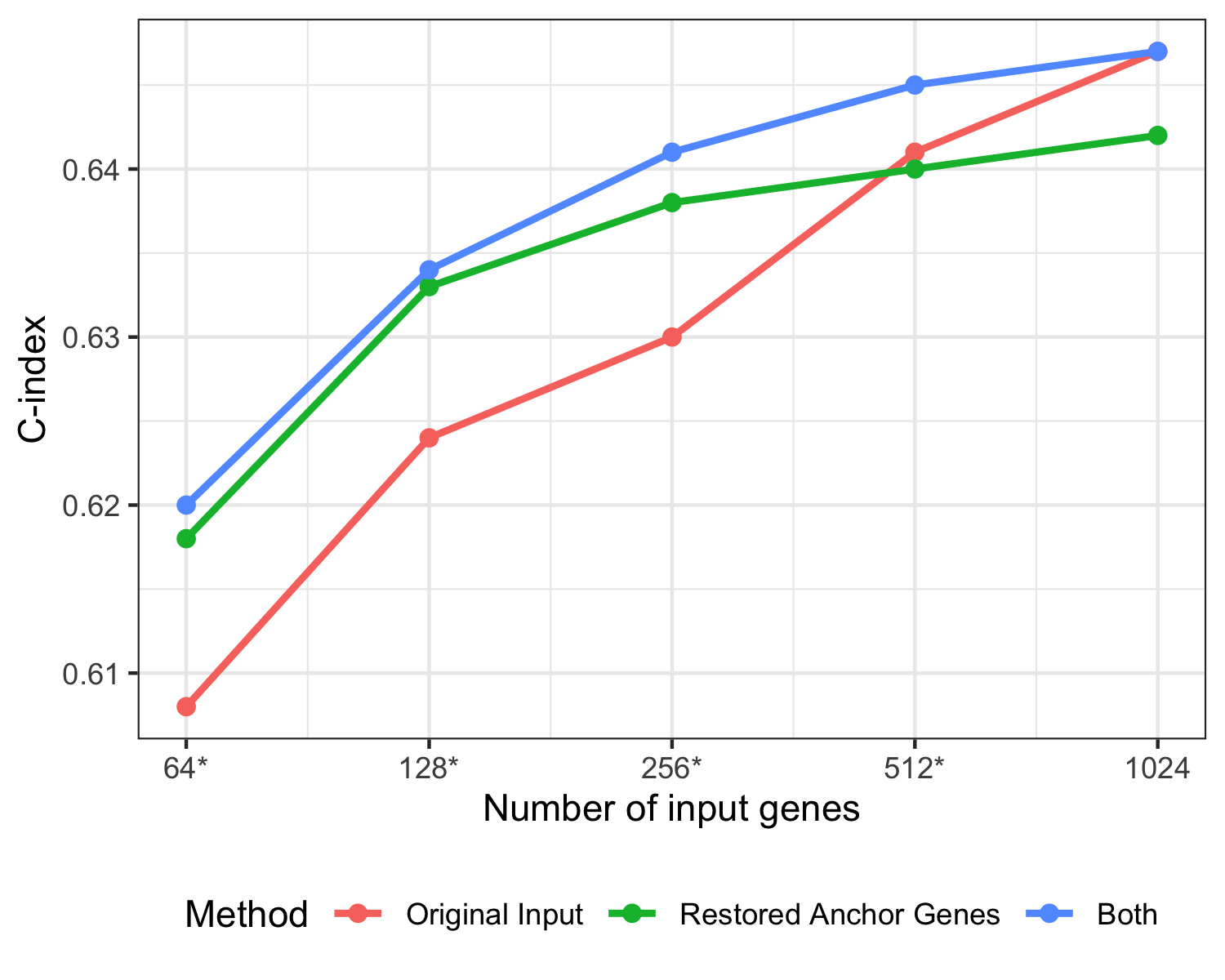}
    \caption{Impact of restored anchor genes on cancer-specific survival prediction. Average concordance index (C-index) across different gene set sizes for three methods: using only original input genes (red), using restored anchor gene expression (green), and combining both (blue). The * symbol indicates statistically significant improvement of the ``Both'' method over the ``Original Input'' method.}
    \label{fig:rna-avg-survival-comp}
\end{figure}

In \autoref{table:rna-survival-pca-cls}, we present the performance evaluation of the summary embedding in cancer-specific survival prediction tasks, comparing it with the PCA approach using the top 200 components. The results show that the summary embedding consistently outperformed PCA and improved as the number of input genes increased. In contrast, PCA performance remained stable regardless of the number of input genes.  

In \autoref{fig:rna-avg-survival-comp}, we compare the performance of cancer-specific survival prediction models built from a random subset of gene expressions. When there are 64 input genes, the CoxPH model with $L_2$ regularization achieved an average C-index of 0.608. As the number of input genes increased, the model’s performance improved, reaching an average C-index of 0.647 with 1,024 input genes.  

With the pre-trained GexBERT model, we encoded the expression information of the selected input genes and used it to restore the expression levels of 512 anchor genes for each cancer type. The restored expression levels were then used for survival prediction. We found that with up to 256 input genes, GexBERT’s restored expression levels outperformed the original expression levels by 1.7\%, 1.4\%, and 1.3\%, respectively. However, with larger input sizes, the performance of GexBERT-restored values lagged behind that of the original expression values. Furthermore, we concatenated the restored expression values with the original input and assessed their impact on survival prediction. The results consistently demonstrated that incorporating the restored expression values of anchor genes improved survival prediction performance compared to using only the original expression levels. This effect was most pronounced with fewer than 256 input genes but diminished when they were more than 512 input genes.

\subsection{Missing data imputation}

\begin{figure}
    \centering
    \includegraphics[width=1.0\textwidth]{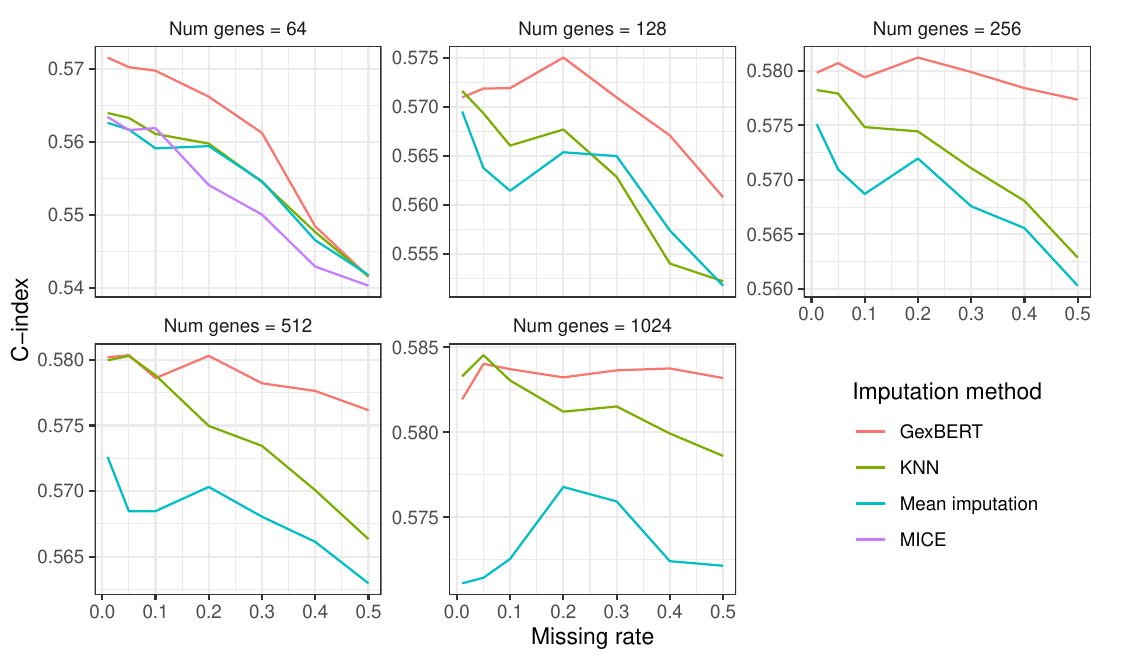}
    \caption{Performance comparison of missing data imputation methods. C-index scores for survival prediction across different missing rates and gene set sizes, using GexBERT (red), KNN (green), mean imputation (blue), and MICE (purple). GexBERT consistently maintains higher predictive performance, demonstrating robustness to missing values compared to traditional imputation methods.}\label{fig:rna-missing-survival-comp}
\end{figure}

The comparison of GexBERT's performance in the presence of missing values with other commonly used missing imputation approaches is summarized in \autoref{fig:rna-missing-survival-comp}. For 64 input genes, GexBERT model outperformed the KNN model, mean imputation approach, and MICE method. Given the high computational cost of MICE and no added benefits comparing to KNN and mean imputation, we excluded MICE as an imputation method for the rest of evaluations. The performance of GexBERT surpasses that of KNN and mean imputation approaches. This underscores the robustness of the pre-trained GexBERT model in handling missing values.

\subsection{Gene embedding}
\begin{figure}
    \centering
    \includegraphics[width=\textwidth]{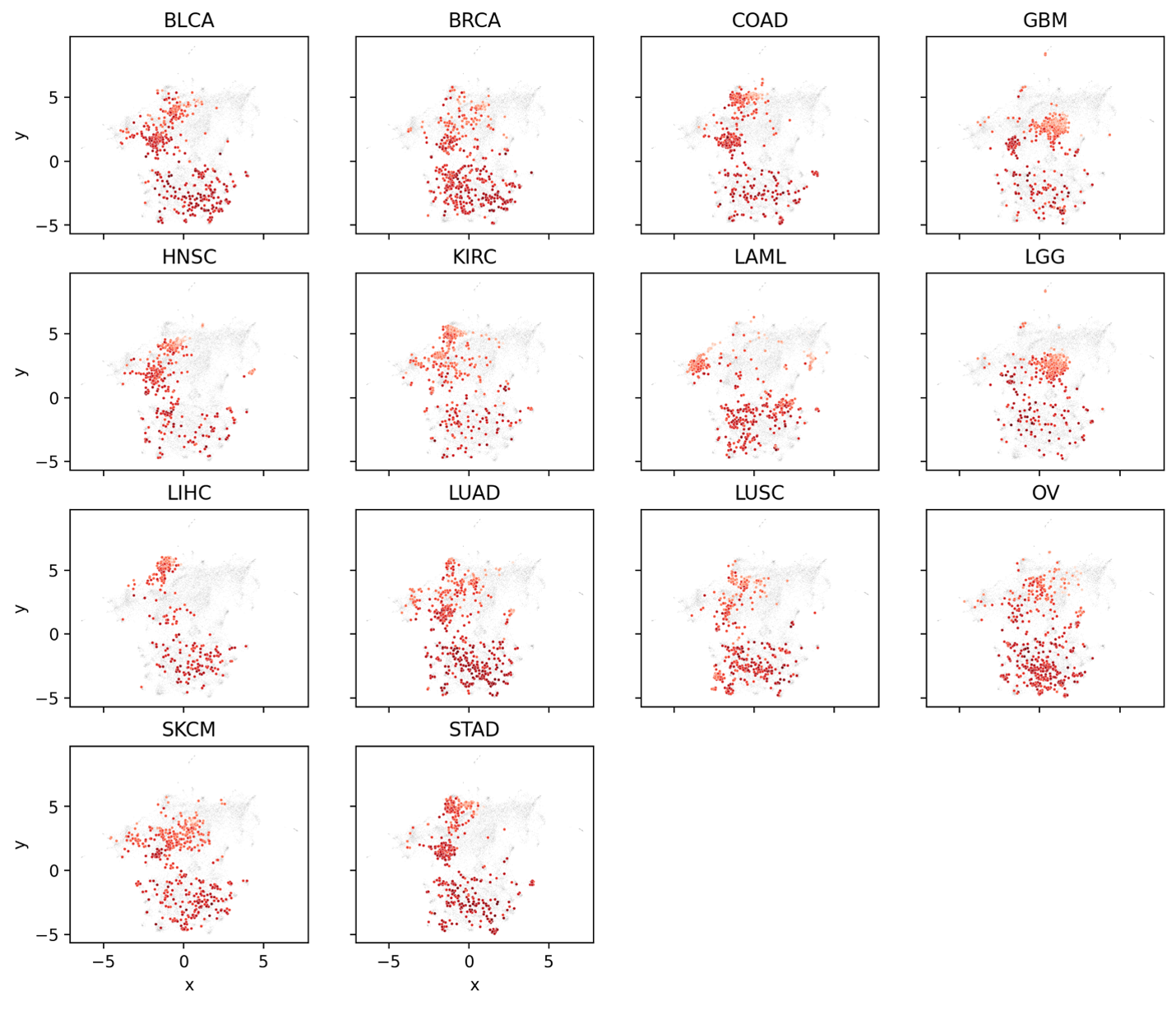}
    \caption{Gene attention patterns across cancer types. UMAP visualization of the top 512 genes with the highest attention weights for each cancer type, as identified by GexBERT. Darker red indicates higher attention values, highlighting key genes that contribute to transcriptomic differentiation.}
    \label{fig:rna-gene-attn-umap}
\end{figure}

% The UMAP visualization\cite{mcinnes2020umap} of gene embeddings learned by GexBERT is presented in Supplementary Figure \autoref{fig:rna-gene-emb-umap} and 
We further explore the structure of the UMAP representation by highlighting the top 512 genes receiving the highest attention weights for each cancer type in \autoref{fig:rna-gene-attn-umap}. The spatial distribution of these highly attended genes varies significantly across cancer types. Notably, cancers with similar etiologies, such as GBM and LGG, display comparable distributions of their top attended genes. This pattern suggests that the attention mechanism of GexBERT captures distinct biological characteristics associated with different cancer types.

\begin{figure}
    \centering
    \includegraphics[width=\textwidth]{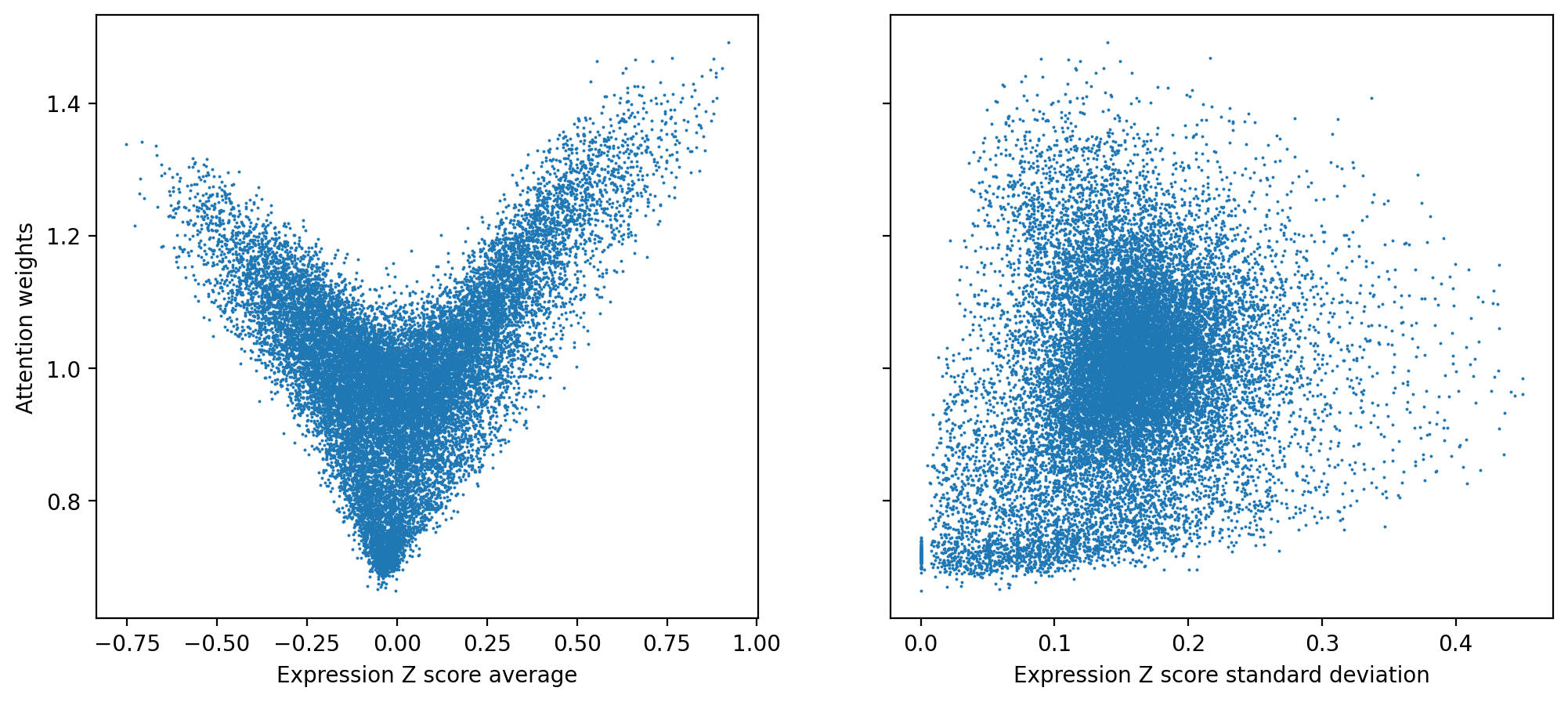}
    \caption{Relationship between GexBERT attention weights and gene expression statistics in LGG.
Scatter plots showing the correlation between attention weights and a) average gene expression Z-scores and b) standard deviation of gene expression Z-scores for LGG cancer. Higher attention weights are assigned to genes with extreme expression values, highlighting their potential biological relevance.}\label{fig:rna-assoc-attn-exp}
\end{figure}

In \autoref{fig:rna-assoc-attn-exp}, we examine the relationship between genes' attention weights and their expression Z-score metrics for a selected cancer type (LGG). The analysis reveals a positive correlation between attention weights and the absolute value of the average expression Z-score. Additionally, genes with greater variability in expression, as indicated by a higher standard deviation of their Z-scores, tend to receive higher attention weights. These findings suggest that GexBERT prioritizes genes exhibiting either differential expression or high variability across patients, highlighting its ability to learn context-aware gene representations.

\subsection{Ablation studies}
\subsubsection{Pre-training strategies}
In this experiment, we evaluated how the selection of genes for restoration might affect the pre-training of the transformer encoder of GexBERT. Our original pre-training used the same set of genes as the input genes for restoration. Subsequently, after the validation loss stopped improving, we further pre-trained this model by using another random subset of genes for restoration. 

\autoref{table:rna-abl-pt} summarizes the results of using the summary embeddings extracted from these two models for pan-cancer classification. The findings reveal that after pre-training the transformer model by selecting a different set of genes for restoration, the summary embeddings exhibited significantly improved performance in pan-cancer classification tasks compared to the model that only restored the same set of genes throughout the pre-training phase. This augmentation might be attributed to the requirement of the transformer encoder to capture higher-level information pertaining to patient characteristics, rather than merely representing the genes present in the input set, when restoring a diverse set of genes.

\begin{table}[]
\centering
\caption{Impact of gene selection strategies during pretraining on classification performance.
Comparison of classification accuracy (mean $\pm$ standard deviation) when GexBERT is pretrained using either a dynamically selected set of restoration genes or the same fixed set across iterations. Selecting different restoration genes improves performance across all gene set sizes, highlighting the benefit of diverse contextual learning.}
\label{table:rna-abl-pt}
\begin{tabular}{@{}lcc@{}}
\toprule
Number of genes & Selecting different set of genes & Using the same set of genes  \\ \midrule
64              & 0.940 (0.011)      & 0.920 (0.011)    \\
128             & 0.962 (0.005)      & 0.951 (0.008)    \\
256             & 0.971 (0.005)      & 0.967 (0.004)    \\
512             & 0.975 (0.003)      & 0.972 (0.003)    \\
1024            & 0.979 (0.002)      & 0.975 (0.002)    \\
% 4096            & 0.983 (0.001)      & 0.978 (0.003)    \\ 
\bottomrule
\end{tabular}
\end{table}

\subsubsection{Number of anchor genes}

\begin{figure}
    \centering
    \includegraphics[width=0.65\textwidth]{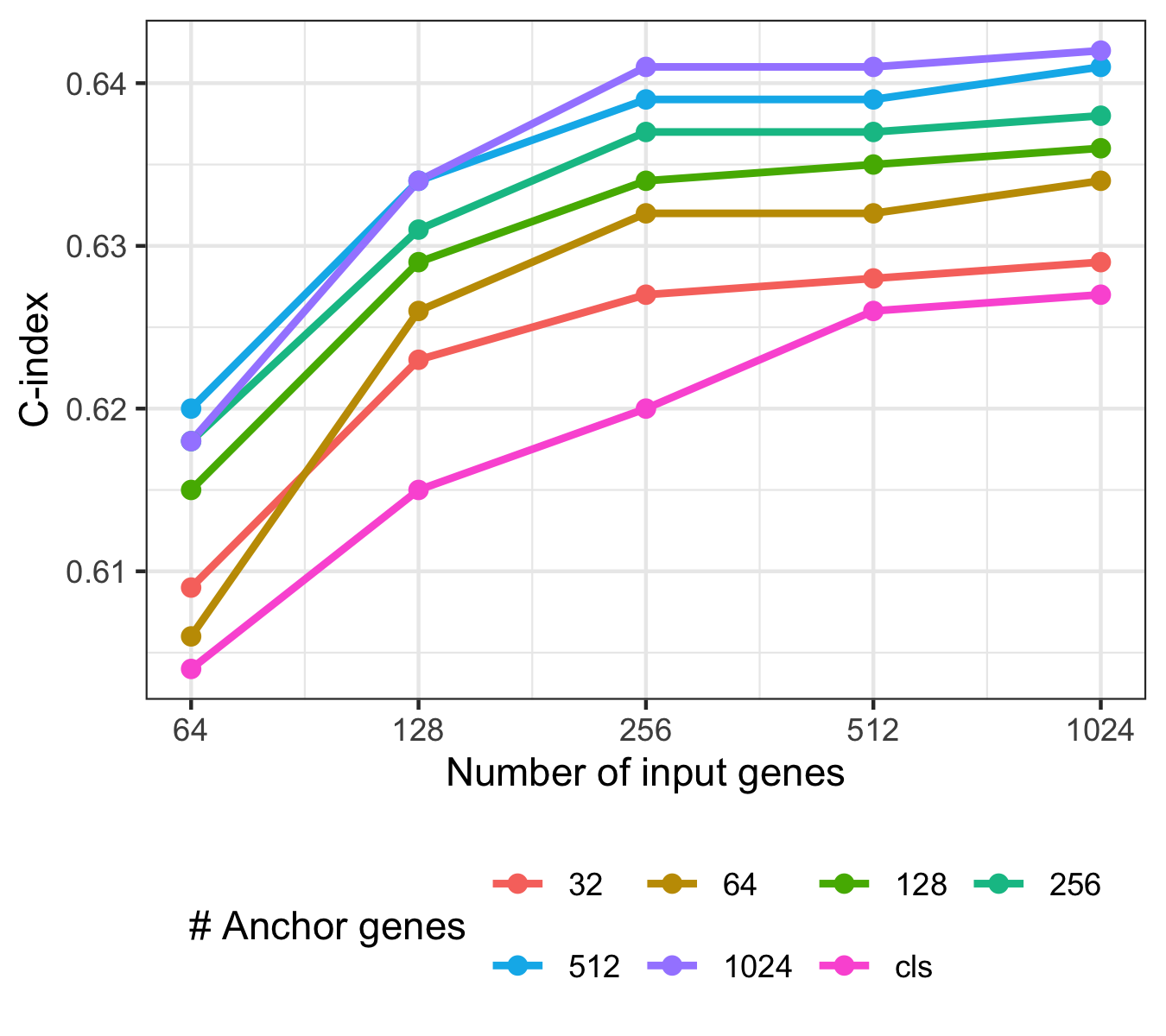}
    \caption{Effect of anchor gene selection on survival prediction performance. Average concordance index (C-index) across different numbers of input genes, comparing survival prediction performance using varying numbers of anchor genes. Larger anchor gene sets improve performance, while the \textit{cls} token-based representation (pink) shows lower predictive power.}\label{fig:rna-abl-survival-anchor}
\end{figure}

We assessed the effect of anchor gene selection on survival prediction performances, and the results are presented in \autoref{fig:rna-abl-survival-anchor}. In general, for any number of input genes, using more anchor genes leads to higher average C-index for survival prediction. Notably, using the summary embedding directly for survival prediction does not yield satisfactory performances compared to using anchor genes, as its performance consistently falls short of models that restore anchor genes.

\section{Discussion}

GexBERT introduces a novel transformer-based framework for learning robust, low-dimensional representations of gene expression data, addressing several long-standing challenges in transcriptomic modeling. A key strength of our approach is its ability to learn contextualized gene embeddings through a pre-training strategy that masks and restores subsets of gene expression values. This design mimics masked language modeling but adapts it to the continuous and high-dimensional nature of gene expression. By capturing co-expression patterns and inter-gene relationships, GexBERT enables accurate phenotype prediction from sparse or partially observed input data—a capability that is particularly valuable in clinical settings where comprehensive profiling is often infeasible.

Our findings demonstrate that GexBERT performs competitively across three principal tasks in cancer genomics: pan-cancer classification, cancer-specific survival prediction, and missing data imputation. In pan-cancer classification, GexBERT consistently outperformed baseline models using only a fraction of the transcriptome. For survival prediction, it achieved higher concordance when using restored anchor gene expressions, highlighting its ability to reconstruct prognostically informative signals. Notably, GexBERT proved more robust to missing data than conventional imputation strategies, maintaining strong predictive performance even at high missingness rates. These results position GexBERT as a scalable and practical alternative for gene expression analysis, with direct applicability to data-limited or cost-constrained scenarios.

An important insight from our work is the biological relevance of the learned attention patterns. The model consistently prioritized genes with high variability or differential expression across patient cohorts, and its attention distributions reflected known cancer-type similarities. This interpretability supports the use of GexBERT not only as a predictive model but also as a hypothesis-generating tool for exploring regulatory programs and biomarker discovery.

There are several limitations to consider. Although our evaluation was conducted on held-out data from TCGA, the pre-training and test sets share common data acquisition and processing pipelines. Future studies should assess GexBERT’s generalization across independently generated datasets, including cross-platform (e.g., microarray to RNA-seq) and cross-population settings. Additionally, we simulated missingness under a Missing Completely At Random (MCAR) assumption. Real-world data may exhibit structured or batch-specific missingness that could influence imputation performance differently.

In conclusion, GexBERT demonstrates that transformer-based representation learning can be effectively adapted to model the transcriptomic landscape. By enabling accurate outcome prediction from partial expression profiles and capturing biologically meaningful gene relationships, our model offers a path toward more scalable, interpretable, and data-efficient approaches in cancer genomics. Its application has the potential to reduce profiling costs, improve prediction accuracy in low-coverage settings, and facilitate integrative analysis in multi-omics research. GexBERT lays the foundation for further exploration of deep contextual models in transcriptomics and precision oncology.

\section{Competing Interests}
The authors declare no competing interests.

\section{Acknowledgement}
This research was supported in part by grants from the US National Library of Medicine (R01LM012837 and R01LM013833) and the US National Cancer Institute (R01CA249758).

\bibliographystyle{unsrt}   %% specify bibliography style (requires IEEEtran.bst somewhere TeX can find it)
\bibliography{8MastersRef}	% expects file "myrefs.bib"

% \begin{suppfigure}
%     \centering
%     \includegraphics[]{gene-emb-umap.png}
%     \caption{Gene Embedding UMAP visualization, colored in normalized average attention weights. In this figure, each gene is color-coded based on its average attention weight in the transformer encoder, relative to the overall average attention weight across all genes. The visualization suggests a pattern where genes positioned in the upper left region of the 2D space tend to have lower attention weights, while those located toward the bottom exhibit higher attention weights.}\label{fig:rna-gene-emb-umap}
% \end{suppfigure}

\end{document}